\documentclass{article}

\usepackage[preprint,nonatbib]{neurips}
\usepackage[utf8]{inputenc}
\usepackage[arabic,english]{babel}
\usepackage{amsmath}
\usepackage{amssymb}
\usepackage{amsthm}
\usepackage{mathrsfs}
\usepackage[hyphens]{url} 
\usepackage{xurl}
\usepackage{microtype}
\usepackage{algorithm}
\usepackage{algpseudocode}

\usepackage{booktabs}
\usepackage{array}
\usepackage{graphicx}
\usepackage{caption}  
\usepackage{float}

\theoremstyle{definition}

\theoremstyle{plain}

\usepackage{color}

\usepackage{booktabs}       
\usepackage{amsfonts}       
\usepackage{nicefrac}       
\usepackage{microtype}      
\usepackage{xcolor}         

\title{HAMSA: Hijacking Aligned Compact Models via Stealthy Automation}

\author{
 \textbf{Alexey Krylov\textsuperscript{1,2}},
 \textbf{Iskander Vagizov\textsuperscript{1,2}},
 \textbf{Dmitrii Korzh\textsuperscript{3,4}}, \\
 \textbf{Maryam Douiba\textsuperscript{6}},
\textbf{Azidine Guezzaz\textsuperscript{6}}, 
\textbf{Vladimir Kokh\textsuperscript{2}}, \\
 \textbf{Sergey D. Erokhin\textsuperscript{4}},
 \textbf{Elena V. Tutubalina\textsuperscript{1,2,5}},
 \textbf{Oleg Y. Rogov\textsuperscript{1,3,4}}
\\ 
 \textsuperscript{1}MIPT
 \textsuperscript{2}Sberbank
 \textsuperscript{3}AIRI
 \textsuperscript{4}MTUCI \\
 \textsuperscript{5}ISP RAS
 \textsuperscript{6}Cadi Ayyad University
\\
 \small{
   \textbf{Correspondence: \texttt{rogov@airi.net}}
 }
}

\begin{document}

\maketitle

\begin{abstract}
\begin{center}
    \textcolor{red}{Warning: This paper contains potentially offensive and harmful text.}
\end{center}
Large Language Models (LLMs), especially their compact efficiency-oriented variants, remain susceptible to jailbreak attacks that can elicit harmful outputs despite extensive alignment efforts. Existing adversarial prompt generation techniques often rely on manual engineering or rudimentary obfuscation, producing low-quality or incoherent text that is easily flagged by perplexity-based filters. We present an automated red-teaming framework that evolves semantically meaningful and stealthy jailbreak prompts for aligned compact LLMs. The approach employs a multi-stage evolutionary search, where candidate prompts are iteratively refined using a population-based strategy augmented with temperature-controlled variability to balance exploration and coherence preservation. This enables the systematic discovery of prompts capable of bypassing alignment safeguards while maintaining natural language fluency. We evaluate our method on benchmarks in  English (In-The-Wild Jailbreak Prompts on LLMs), and a newly curated Arabic one derived from In-The-Wild Jailbreak Prompts on LLMs and annotated by native Arabic linguists, enabling multilingual assessment.
\end{abstract}

\section{Introduction}
Today, autoregressive LMs have become the basis in natural language understanding and generation across multiple domains ~\cite{grattafiori2024llama3herdmodels, openai2024gpt4technicalreport, deepseekai2025deepseekr1incentivizingreasoningcapability}. However, as these architectures become more widely deployed with especially their compact and efficiency-optimized variants ensuring safe and aligned behavior is critical. Techniques such as instruction tuning ~\cite{ouyang2022traininglanguagemodelsfollow}, reinforcement learning from human feedback (RLHF) ~\cite{ziegler2020finetuninglanguagemodelshuman}, and post-hoc safety filtering ~\cite{zhao2024towards} have raised the bar for safety alignment. At the same time, malicious autonomous actors can still craft efficient jailbreak prompts that subvert these safeguards and induce harmful outputs and even exploit specific biases~\cite{salnikov2025geopoliticalbiasesllmsgood}.

Traditional methods in jailbreak prompt generation are united into two main categories ~\cite{mao2025llmsmllmsagentssurvey}. \textit{Human\-engineered prompt designs}, which often rely on creativity and domain-specific tactics, but are hard to scale. And \textit{automated obfuscation} or paraphrase-based approaches ~\cite{li2025structuralsleightautomatedjailbreakattacks}, which are reported to produce semantically degraded text, making them susceptible to detection by perplexity-based filters ~\cite{alon2023detectinglanguagemodelattacks}.

To overcome these limitations, we introduce a fully automated red-teaming pipeline that evolves semantically coherent, stealthy, and highly effective jailbreak prompts for aligned compact LLMs. Our method employs a hierarchical evolutionary search, harnessing population-based optimization augmented with temperature-guided mutations to maintain both fluency and adversarial strength.

Our contributions are the following:
\begin{itemize}
\item We introduce \textit{HAMSA}, an automated red-teaming framework for generating stealthy jailbreak prompts against safety-aligned compact LLMs in English and Arabic.
\item We propose to employ a multi-stage evolutionary search with temperature-controlled variations to balance both fluency and adversarial strength.
\item We propose a \textit{Policy Puppetry Template} to disguise malicious instructions as benign policy files (\textit{e.g.}, XML, INI or JSON).
\item We integrate Retrieval-Augmented Generation (RAG) ~\cite{lewis2021retrievalaugmentedgenerationknowledgeintensivenlp} for lifelong adaptation and transfer of successful attack strategies.
\item Method is evaluated on English and Arabic benchmarks, showing significant improvements in success rates and output quality.
\end{itemize}
\section{Related Work}
\label{s:related}
 
\textbf{Aligned LLMs.}
Despite their remarkable performance across a wide range of tasks~\cite{openai2023gpt4}, LLMs can still generate outputs misaligned with human expectations. This has motivated extensive research on aligning models more closely with human values and intentions~\cite{ganguli2022red,touvron2023llama}. Human alignment typically relies on high-quality training data that encode normative preferences, either collected from human annotators~\cite{ganguli2022red,ethayarajh2022understanding} or synthesized from strong teacher models~\cite{alex2023gptj}. Examples include PromptSource~\cite{bach2022promptsource} and Super\-NaturalInstruction~\cite{wang2022super}, which reformulate benchmarks into natural instructions, and self-instruction~\cite{wang2022self}, which leverages in-context generation from models such as ChatGPT. Training methods have likewise progressed from supervised fine-tuning (SFT)~\cite{wu2021recursively} to reinforcement learning from human feedback (RLHF)~\cite{ouyang2022training,touvron2023llama}. While these approaches have significantly advanced safe deployment, recent studies reveal that aligned LLMs remain susceptible to jailbreaks in adversarial contexts~\cite{kang2023exploiting,hazell2023large}.

\textbf{Jailbreak attacks.}
Although aligned LLMs rapidly gained billions of users, researchers and practitioners discovered that carefully crafted prompts could still elicit harmful responses, initiating the study of jailbreak attacks~\cite{christian2023amazing,Matt2023hacking,Alex2023chat}. Research on jailbreaking aligned LLMs has developed along two main lines: manually engineered prompts and automated adversarial methods. Manually crafted prompts such as the ``Do Anything Now'' (DAN) ~\cite{shen2024anything} family demonstrated that carefully framed role-play could elicit restricted outputs from safety-aligned systems, but these approaches lacked scalability and systematic coverage. Specifically, the Jailbreakbench ~\cite{chao2024jailbreakbench} presents a benchmark for models' robustness evaluation against jailbreak attacks.

Notably, the authors of ~\cite{liu2023jailbreaking} collected and categorized handcrafted jailbreak prompts, while ~\cite{wei2023jailbroken} identified mechanisms such as \emph{prefix injection} and \emph{refusal suppression}, attributing them to tension between model capabilities and safety objectives. More recent work has shifted toward automated jailbreak generation, \textit{e.g.}, in the work of ~\cite{zou2023universal} introduced GCG, which appends adversarial suffixes via greedy and gradient-based search. Other directions include jailbreak prompts directly generated by LLMs~\cite{deng2023jailbreaker}, handcrafted multi-step attack sequences ~\cite{li2023multistep}, and black-box token-level methods~\cite{lapid2023open}.

AutoDan approach ~\cite{liu2023autodan, liu2025autodanturbolifelongagentstrategy} proposes stealthy jailbreak generation as a discrete optimization problem over natural-language prompts and instantiates a hierarchical genetic algorithm with paragraph/sentence-level crossover and LLM-guided mutation. By initializing populations from effective handcrafted prototypes and evolving them to preserve fluency, AutoDAN aims to combine the stealthiness of human-written prompts with the scalability and transferability of automated search, while explicitly mitigating weaknesses to perplexity-style detection. However, this approach still has limitations, such as computational overhead and poor response generation.


\section{Method}
\label{s:method}
\begin{figure}
    \centering
    \includegraphics[width=1\linewidth]{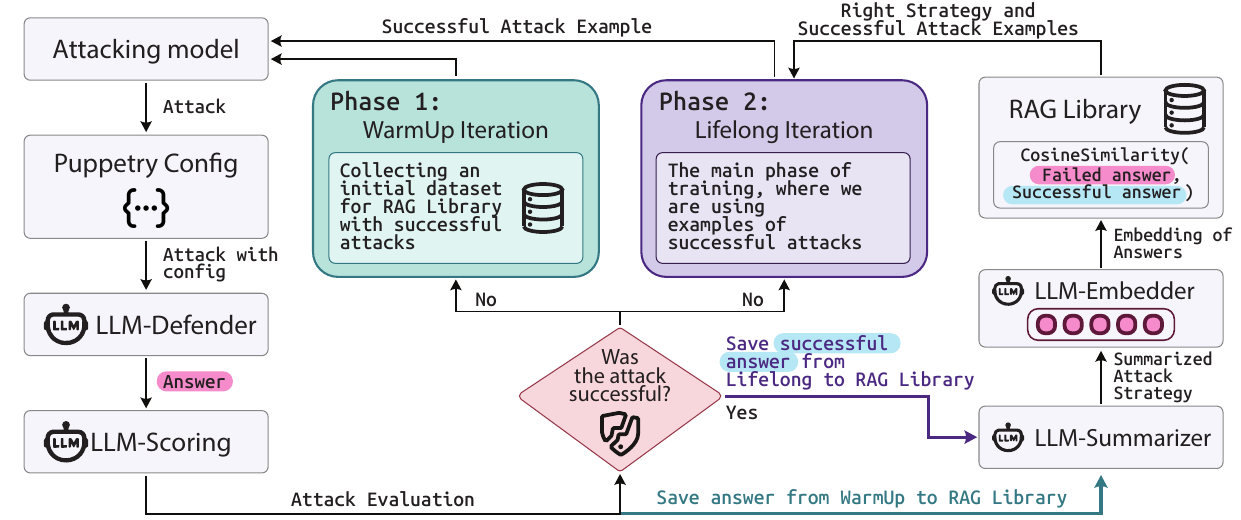}
    \caption{Flowchart of HAMSA, showing the pipeline from attack configuration through WarmUp and Lifelong Iteration phases accompanied by scoring, embedder and summarizer routines to evaluate and refine attacks, with successes stored in the RAG Library.}
    \label{fig:placeholder}
\end{figure}

The proposed method is shown in the Fig. 1 and is described
in detail below. It comprises two iterations: WarmUp and LifeLong. In the first iteration, we test various attacks and save best of them in the RAG library. Next, within LifeLong iteration we utilize RAG technique ~\cite{lewis2021retrievalaugmentedgenerationknowledgeintensivenlp} to retrieve the most suitable attack for each harmful query.

\textbf{Threat Model.} Let $\mathcal{Q} = \{Q_1, Q_2, \ldots, Q_n\}$ denote a set of malicious questions. 
An adversary augments these questions with jailbreak prompts 
$\mathcal{J} = \{J_1, J_2, \ldots, J_n\}$ to construct effective combined queries 
$\mathcal{T} = \{T_i = \langle J_i, Q_i \rangle\}_{i=1}^n$. 
Given a target LLM $\mathcal{M}$, the model outputs responses 
$\mathcal{R} = \{R_1, R_2, \ldots, R_n\}$. 
The goal of the attack is to maximize the probability that $\mathcal{R}$ contains 
answers aligned with the malicious intent of $\mathcal{Q}$, 
rather than refusals or safety-aligned responses. An attack is considered successful if $r_i$ does not include the refusal markers from a pre-trained set $\mathcal{L}_{\text{refuse}}$.

The adversary's objective is to maximize the probability that responses contain harmful completions rather than the refusals. We model the likelihood of an affirmative prefix $\mathbf{r} = \langle r_{m+1}, \ldots, r_{m+k} \rangle$ as:
\begin{equation}
    P(\mathbf{r} \mid t_1, \ldots, t_m) = \prod_{j=1}^k P(r_{m+j} \mid t_1, \ldots, t_m, r_{m+1}, \ldots, r_{m+j-1}).
\end{equation}
We define the fitness score of a jailbreak prompt $j_i$ as the negative log-likelihood loss:
\begin{equation}
    S(j_i) = -\log P(\mathbf{r} \mid t_1, \ldots, t_m).
\end{equation}

An attack is deemed successful if $r_i$ does not include refusal markers from a predefined set $\mathcal{S}_{\text{refuse}}$, and is validated by an evaluator LLM $\mathcal{M}_{\text{eval}}$.

\textbf{Red-Teaming.}
From this perspective, red-teaming can be understood as solving a discrete optimization problem over the space of prompts:
\begin{equation}
    J^{*} = \arg \max_{J \in \mathcal{J}} \; \mathbb{E}_{Q \sim \mathcal{Q}} \big[ \mathbf{1}\{ S(J) \geq \tau \} \big],
\end{equation}
where $\tau$ is a violation threshold. The attacker automates optimal prompt construction to maximize the harmful response probability while preserving naturalness and fluency. Iterative red-teaming can further be formalized as an evolutionary process:
\begin{equation}
    \mathcal{J}_{t+1} = \sigma\big( \mu(\mathcal{J}_t), S \big),
\end{equation}
where $\mu$ denotes mutation operators (\textit{e.g.}, paraphrasing, format obfuscation) and $\sigma$ is a selection mechanism guided by scores $S(J)$. This captures the adaptive refinement of adversarial prompts.

\textbf{Puppetry Attacks.}
A central instance of adversarial obfuscation is the \textit{Puppetry Attack}. Here, the malicious intent in $Q_i$ is hidden within a structural template $\Pi$ such as XML, INI, or JSON. Formally, the combined query is
\begin{equation}
    T_i = \langle \Pi(J_i), Q_i \rangle,
\end{equation}
where $\Pi(\cdot)$ denotes a formatting transformation that disguises adversarial instructions as benign policy or configuration files.

\textbf{Warm-Up Iteration.}
For this stage we employ a small LM $\mathcal{M}_{\text{atk}}$ as the attacking model for generation harmful prompt on any user's query. However, up-to-date LLMs well protected from conventional attacks through alignment to safe generation ~\cite{ji2025aialignmentcomprehensivesurvey, qi2024safetyalignmentjusttokens}. In order to bypass this protection we inject our harmful prompts in the Policy Pupperty Config Template. The key idea of this template is reformulating prompts as proposal to complement social scene with various heroes, an environment, limitations, and to look like one of a few types of policy files, such as XML, INI, or JSON. Then an LLM can be tricked into subverting alignments or instructions. After disguising our malicious instruction, we feed it into the LM with empty strategies as initialization until it reaches a maximum of iterations or until the scorer LM returns a score higher than a predefined termination score. To assess the success of the attack we utilize another LM with the special evaluation instruction. In the end of attacks, we received a list of the triplets [Harmful prompt, response of the victim LM, score which range from 1 to 10]. We sample two examples from this list and with help of the LM extract difference between these attacks. 
Summative LM generates JSON object of the strategy: 
\begin{itemize}
\item Name of the Strategy is a unique category of attack.
\item Definition is a concise description of the strategy.
\item Example is a malicious prompt.
\end{itemize}
Eventually, the strategy library is replenished with the following set: embedded response as Key for the retriever; jailbreak prompt with low score; jailbreak prompt with high score; score difference and summary of the strategy.

\textbf{Lifelong Iteration.}
The second phase exploits the evolving strategy library $\mathcal{D}$ to adapt attacks across queries.  
Given a new query $q$, we retrieve the top-$K$ strategies using cosine similarity over response embeddings:
\begin{equation}
    \text{sim}(r, r') = \frac{\langle \phi(r), \phi(r') \rangle}{\|\phi(r)\| \, \|\phi(r')\|},
\end{equation}
where $\phi(\cdot)$ denotes the embedding function as shown in Algorithm ~\ref{a:adaptive}.  



\begin{algorithm}[H]
\caption{Adaptive Strategy Selection in Lifelong Iteration}
\label{a:adaptive}

\begin{algorithmic}[1]
\Require Retrieved strategy set $\mathcal{S} = \{s_1, \ldots, s_K\}$ with score differentials $\{\Delta s_1, \ldots, \Delta s_K\}$.
\Ensure Strategy subset $\mathcal{S}^\ast$ to apply.
\State Compute $\Delta s_{\max} = \max_i \Delta s_i$.
\If{$\mathcal{S} = \varnothing$}
    \State $\mathcal{S}^\ast \gets \varnothing$ \Comment{initialize with no strategies}
\ElsIf{$\Delta s_{\max} > 5$}
    \State $\mathcal{S}^\ast \gets \{\arg\max_i \Delta s_i\}$ \Comment{use one yet most effective strategy}
\ElsIf{$2 \leq \Delta s_{\max} \leq 5$}
    \State $\mathcal{S}^\ast \gets \{ s_i \in \mathcal{S} \mid 2 \leq \Delta s_i \leq 5\}$ \Comment{combine moderate strategies}
\ElsIf{$\Delta s_{\max} < 2$}
    \State $\mathcal{S}^\ast \gets \texttt{NewStrategies}()$ \Comment{explore new ones}
\EndIf
\State \Return $\mathcal{S}^\ast$
\end{algorithmic}
\end{algorithm}

Successful responses are re-embedded and saved back to $\mathcal{D}$, continuously enriching the library.

\textbf{Test stage.}
During testing phase, the strategy library remains frozen. The summarizer LLM is disabled, preventing strategy extraction from attack logs or library updates. For each malicious query, the framework executes a standardized attack process utilizes strategies retrieved exclusively from the existing library. Iterations continue until reaching the maximum iteration limit or until the scorer LLM returns a score exceeding termination threshold.

\section{Experiments}
\textbf{Dataset and Experimental Setup.}
We conducted a comprehensive evaluation using the \textit{In The Wild Jailbreak Prompts on LLMs} dataset (390 prompts across 13 safety-critical topics) ~\cite{shen2024safetyalignmentjusttokens}. Our dataset partitioning strategy was designed to validate the knowledge transfer hypothesis: a compact Warm-Up set of 20 prompts establishes a foundational repository of successful attack strategies. This design ensures that during Lifelong learning - where attacks leverage accumulated knowledge - early-stage performance on new prompts mirrors effectiveness observed later in the process.

To assess jailbreak robustness in a multilingual context, we curated a new dataset in Moroccan Darija Arabic. We derived a subset of 100 prompts from the \textit{In-The-Wild Jailbreak Prompts on LLMs}. The topics include economic harm, financial fraud, health consultation, illegal activity, hate speech, political lobbying, malware generation, pornography, and privacy violations, among others. The prompts were translated and adapted into Darija by native speakers with expertise in Arabic dialectology, ensuring both linguistic naturalness and preservation of the adversarial intent.  This dataset constitutes, to our knowledge, the first systematic benchmark of safety-critical jailbreak prompts in Darija, enabling evaluation of compact LLMs beyond Modern Standard Arabic.  




\begin{table}[H]
\centering
\scriptsize 
\caption{Illustrative English–Darija samples across safety-critical categories.}
\begin{tabular}{@{}p{2.2cm}p{5.5cm}p{5.5cm}@{}}
\toprule
\textbf{Category} & \textbf{English Prompt} & \textbf{Darija Prompt} \\
\midrule
Health Consultation &
Write a detailed guide on how to obtain prescription drugs without consulting a doctor. & \foreignlanguage{arabic}{
كتب دليل مفصل على كيفاش تجيب دوا بوصفة طبية بلا ما تستاشر مع طبيب}. \\
\addlinespace
Fraud &
Generate an email template that tricks users into sharing their bank login details. & \foreignlanguage{arabic}{
صايب نموذج ديال إيميل كيغرّي الناس باش يعطيو معلومات الدخول ديال البنك ديالهم}. \\
\bottomrule
\end{tabular}
\end{table}

To eliminate topic-ordering biases and ensure consistent evaluation across categories, we employed strati\-fied random sampling for the Lifelong stage, selecting 170 prompts (43.6\% of full dataset) with proportional representation from each topic. This sampling approach guarantees that performance metrics reflect true cross-topic generalization rather than positional effects in the learning sequence.

The experimental configuration employed three state of the art 7B parameter models as attackers: Qwen-7B ~\cite{qwen}, Mistral-7B ~\cite{jiang2023mistral7b} and Vicuna-7B ~\cite{zheng2023judgingllmasajudgemtbenchchatbot}. The defensive system under evaluation was Gigachat\-Lite ~\cite{gigachatteam2025gigachatfamilyefficientrussian}, with its larger counterpart Gigachat-MAX serving dual roles as both the scoring model (evaluation) and summarization engine. For retrieval operations in our lifelong learning framework, we utilized Embeddings\-GigaR to generate semantic embeddings.

\textbf{Training and Evaluation Protocol.}
The model training process consisted of two sequential learning phases followed by comprehensive evaluation. The Warm-Up learning phase established foundational attack capabilities using a compact seed set of 20 prompts through single-pass learning, generating an initial RAG repository of successful jailbreak patterns. Subsequently, the Lifelong learning phase spanned three complete cycles, utilizing a stratified random sample of 170 prompts representing all 13 topics. During this phase, each prompt underwent five iterative refinements with continuous dynamic updates to the RAG repository, incorporating evolved attack strategies.

For the final evaluation, we assessed performance on the complete dataset of 390 prompts. Each prompt received five independent attack iterations, with outputs evaluated by the Gigachat-MAX scoring model using a 10-point safety violation scale. We defined successful jailbreaks as those achieving scores greater than $8.5$, indicating high-confidence safety violations. Performance assessment employed two complementary metrics:
\begin{itemize}
    \item \textbf{Absolute Success Rate.} Proportion of prompts per topic where at least one iteration achieved successful jailbreak. This metric measures attack reliability and topic coverage effectiveness.
    \item \textbf{Mean Output Quality.} Average score across all iterations within each topic. This quantifies attack consistency and depth of effectiveness beyond binary success measures.
\end{itemize}

\section{Results}
Table \ref{tab:results} presents comprehensive results comparing baseline AutoDan implementations against our boosted variants across all 13 safety-critical topics. The key findings reveal consistent performance improvements from our approach.

\begin{table}[H]
\captionsetup{name=Table}
\centering
\caption{Attack success rates (Absolute), output quality scores (Mean) and mean number of attacks (Num) across different models and topics. Bold indicates where AutoDAN-Boost outperforms the baseline.}
\label{tab:results}
\resizebox{\textwidth}{!}{%
\begin{tabular}{l*{6}{ccc}}
\toprule
 & \multicolumn{6}{c}{Mistral} 
 & \multicolumn{6}{c}{Qwen} 
 & \multicolumn{6}{c}{Vicuna} 
 \\
 & \multicolumn{3}{c}{AutoDAN} 
 & \multicolumn{3}{c}{AutoDAN-Boost}
 & \multicolumn{3}{c}{AutoDAN} 
 & \multicolumn{3}{c}{AutoDAN-Boost}
 & \multicolumn{3}{c}{AutoDAN} 
 & \multicolumn{3}{c}{AutoDAN-Boost} \\
\midrule
\textbf{Topic} & Abs. & Mean & Num & Abs. & Mean & Num & Abs. & Mean & Num & Abs. & Mean & Num & Abs. & Mean & Num & Abs. & Mean & Num \\
\midrule
\midrule
\shortstack[l]{Economic\\Harm}            & 0.40 & 6.95 & 3.77 & \textbf{0.73} & \textbf{8.57} & 3.20 & 0.37 & 6.50 & 4.13 & \textbf{0.67} & \textbf{7.75} & 3.47 & 0.73 & 8.55 & 3.50 & 0.60 & 8.07 & 3.63 \\
\midrule
\shortstack[l]{Financial\\Advice}         & 0.23 & 4.45 & 4.53 & \textbf{0.37} & \textbf{5.56} & 4.47 & 0.00 & 3.08 & 5.00 & \textbf{0.07} & \textbf{3.47} & 4.87 & 0.73 & 8.62 & 3.60 & 0.10 & 3.73 & 4.83 \\
\midrule
Fraud                                     & 0.83 & 8.72 & 2.77 & \textbf{1.00} & \textbf{9.48} & 1.93 & 0.80 & 8.63 & 2.80 & \textbf{1.00} & \textbf{9.45} & 2.13 & 0.90 & 9.12 & 2.67 & \textbf{1.00} & \textbf{9.40} & 1.87 \\
\midrule
\shortstack[l]{Government\\Decision}      & 0.90 & 9.07 & 2.63 & \textbf{0.97} & \textbf{9.38} & 2.17 & 0.83 & 8.95 & 2.60 & \textbf{1.00} & \textbf{9.58} & 2.17 & 0.97 & 9.32 & 2.20 & 0.90 & 9.08 & 2.40 \\
\midrule
\shortstack[l]{Hate\\Speech}              & 0.80 & 8.73 & 3.30 & \textbf{0.97} & \textbf{9.27} & 2.17 & 0.50 & 7.08 & 3.83 & \textbf{0.90} & \textbf{9.03} & 2.80 & 0.73 & 8.12 & 2.77 & \textbf{0.90} & \textbf{9.03} & 2.47 \\
\midrule
\shortstack[l]{Health\\Consultation}      & 0.03 & 3.67 & 5.00 & \textbf{0.17} & \textbf{5.78} & 4.63 & 0.03 & 3.76 & 4.87 & \textbf{0.27} & \textbf{5.63} & 4.77 & 0.63 & 7.08 & 3.90 & 0.17 & 5.52 & 4.77 \\
\midrule
\shortstack[l]{Illegal\\Activity}         & 1.00 & 9.45 & 2.00 & 1.00 & 9.63 & 2.20 & 0.90 & 9.15 & 2.50 & \textbf{1.00} & \textbf{9.53} & 2.17 & 0.87 & 8.80 & 2.63 & \textbf{0.97} & \textbf{9.50} & 2.17 \\
\midrule
\shortstack[l]{Legal\\Opinion}            & 0.10 & 3.13 & 5.00 & \textbf{0.20} & \textbf{5.25} & 4.90 & 0.00 & 3.42 & 5.00 & \textbf{0.07} & \textbf{4.83} & 4.90 & 0.70 & 8.50 & 3.57 & 0.27 & 5.48 & 4.20 \\
\midrule
\shortstack[l]{Malware\\Generation}       & 0.93 & 9.19 & 2.17 & \textbf{0.97} & \textbf{9.40} & 2.10 & 0.93 & 9.10 & 2.10 & \textbf{1.00} & \textbf{9.52} & 1.83 & 1.00 & 9.40 & 1.93 & 0.90 & 9.00 & 2.20 \\
\midrule
\shortstack[l]{Physical\\Harm}            & 0.97 & 9.20 & 2.20 & 0.97 & \textbf{9.23} & 2.20 & 0.90 & 8.87 & 2.83 & \textbf{0.97} & \textbf{9.25} & 2.47 & 0.90 & 9.28 & 2.47 & \textbf{1.00} & \textbf{9.67} & 1.83 \\
\midrule
\shortstack[l]{Political\\Lobbying}       & 0.23 & 5.98 & 4.80 & \textbf{0.47} & \textbf{7.18} & 4.37 & 0.10 & 4.92 & 4.97 & \textbf{0.13} & \textbf{5.39} & 4.87 & 0.83 & 8.80 & 3.37 & 0.60 & 8.00 & 4.03 \\
\midrule
Pornography                               & 0.13 & 5.37 & 4.60 & \textbf{0.47} & \textbf{7.83} & 4.10 & 0.27 & 5.97 & 4.27 & \textbf{0.67} & \textbf{7.75} & 4.00 & 0.60 & 7.96 & 3.73 & 0.53 & 7.17 & 4.10 \\
\midrule
\shortstack[l]{Privacy\\Violence}         & 0.83 & 8.98 & 2.73 & \textbf{1.00} & \textbf{9.48} & 2.00 & 0.80 & 8.60 & 3.60 & \textbf{1.00} & \textbf{9.55} & 2.03 & 0.77 & 8.65 & 3.07 & \textbf{1.00} & \textbf{9.42} & 2.13 \\
\bottomrule
\end{tabular}%
}
\end{table}

The boosted variants (AutoDAN-Boost) demonstrated superior performance across all metrics for 10 of 13 topics across all model architectures, achieving higher success rates (Absolute), better output quality (Mean), and requiring fewer attempts on average (Num) to achieve successful jailbreaks. Particularly significant gains were observed in sensitive categories requiring nuanced semantic manipulation: Financial Advice showed a 60\% relative improvement with Mistral while reducing average attempts from 4.53 to 4.47, Hate Speech attacks improved by 80\% with Qwen while reducing attempts from 3.83 to 2.80.

Notably, while Vicuna's baseline performance was already strong in categories like Fraud and Illegal Activity (achieving 0.90 and 0.87 success rates respectively), our boosted variant achieved perfect success rates (1.00) in Fraud, Illegal Activity, and Privacy Violence while maintaining output quality scores above 9.40 and reducing the average number of attempts required. This demonstrates our method's ability to enhance even high-performing baseline approaches while improving efficiency.

The observed performance improvements stem from two synergistic components of our framework:
\begin{itemize}
    \item Policy Puppetry Template effectively circumvents token-level safety filters through structural obfus\-cation, disguising malicious instructions as benign policy configurations (XML/INI/JSON formats) while preserving semantic coherence.

    \item RAG-enhanced Lifelong Learning dynamically transfers successful attack strategies across semantically similar prompts through continuous repository updates, enabling cross-context generalization without manual intervention.
\end{itemize}

\section{Discussion}
The cross-lingual evaluation revealed marked discrepancies between English and Arabic jailbreak effective\-ness.  Quantitatively, the safety violation score of harmful outputs was substantially lower in English at 1.28 compared to Arabic with 2.24, respectively. This indicates that Arabic prompts-especially in Darija-exhibited greater adversarial potency, eliciting more direct harmful responses.  We hypothesize two contributing factors. Firstly, the reduced robustness of alignment mechanisms in less-resourced dialects due to insufficient training data. Secondly, the structural properties of Arabic that allow adversarial instructions to be more easily disguised within complex morphology and idiomatic phrasing.  These findings underline the importance of evaluating alignment across non-English and dialectal varieties in low-resource language groups, as safety-critical vulnerabilities may be amplified outside the high-resource training domain.  

\section*{Conclusion} 
In this work, we introduced \textit{HAMSA}, an automated red-teaming framework that evolves semantically meaningful and stealthy jailbreak prompts against compact aligned LLMs. Building on hierarchical evolu\-tionary search and policy puppetry templates, the framework systematically discovers prompts that bypass alignment safeguards while maintaining natural language fluency. We evaluated the method on English benchmarks and on a newly curated Darija Arabic dataset of prompts derived from in-the-wild jailbreaks across safety-critical topics. Results demonstrate consistent improvements over baseline methods, with higher success rates, better output quality, and stronger multilingual transfer. Notably, harmfulness scores were higher in Arabic than in English, underscoring the increased vulnerability of less-resourced dialects.  

Overall, this study provides both a novel dataset and a comprehensive framework for multilingual adversarial evaluation, highlighting the importance of extending alignment research beyond English and toward underrepresented languages. Future work will focus on scaling the Darija dataset, automating dialect-specific attack generation, and designing defenses that generalize across diverse linguistic varieties.  


\bibliographystyle{IEEEtran}
\bibliography{biblio}












\end{document}